\documentclass[11pt,a4paper]{article}
\usepackage{verbatim}
\usepackage[hyperref]{emnlp2018}
\usepackage{times}
\usepackage{latexsym}
\usepackage{amsmath}
\usepackage{amssymb}
\usepackage{float}
\usepackage{placeins}
\usepackage[OT1]{fontenc}
\usepackage{graphicx}
\usepackage{algorithm}
\usepackage{hyperref}
\usepackage[noend]{algpseudocode}
\usepackage[small,compact]{titlesec}
\usepackage[font=small,aboveskip=4pt,belowskip=0pt]{caption}
\usepackage{enumitem}
\usepackage[nopar]{lipsum}
\setdescription{noitemsep,topsep=0pt}
\graphicspath{ {images/} }

\DeclareMathOperator{\softmax}{softmax}
\DeclareMathOperator{\enc}{enc}
\DeclareMathOperator{\dec}{dec}
\DeclareMathOperator{\precedes}{precedes}
\DeclareMathOperator{\loss}{Loss}
\DeclareMathOperator{\nan}{Null}
\DeclareMathOperator{\labels}{label}
\DeclareMathOperator{\topo}{topo}
\DeclareMathOperator{\pr}{P}
\makeatletter
\def\BState{\State\hskip-\ALG@thistlm}
\makeatother
\algnewcommand{\LeftComment}[1]{\Statex \(\triangleright\) #1}

\usepackage{url}

\newif\ifLossCurve
\newif\ifParseIWSLT
\LossCurvefalse
\ParseIWSLTtrue

\aclfinalcopy 

\title{Top-down Tree Structured Decoding with Syntactic Connections for Neural
       Machine Translation and Parsing}

\author{Jetic G\=u, ~ Hassan S. Shavarani, ~ Anoop Sarkar\\
    TASC 9408, 8888 University Drive, \\
    Simon Fraser University,\\
    Burnaby, BC V5A 1S6, Canada \\
    {\tt \{jeticg, sshavara, anoop\}@sfu.ca}
}

\date{}

\begin{document}
\maketitle
\setlength{\abovedisplayskip}{0pt}%
\setlength{\belowdisplayskip}{0pt}%
\setlength{\abovedisplayshortskip}{0pt}%
\setlength{\belowdisplayshortskip}{0pt}%
\setlength{\jot}{0pt}

\begin{abstract}
  The addition of syntax-aware decoding in Neural Machine Translation (NMT)
  systems requires an effective tree-structured neural network, a syntax-aware
  attention model and a language generation model that is sensitive to sentence
  structure.
  We exploit a top-down tree-structured model called DRNN (Doubly-Recurrent
  Neural Networks) first proposed by Alvarez-Melis and Jaakola (2017) to create
  an NMT model called Seq2DRNN that combines a sequential encoder with
  tree-structured decoding augmented with a syntax-aware attention model.
  Unlike previous approaches to syntax-based NMT which use dependency parsing
  models our method uses constituency parsing which we argue provides useful
  information for translation.
  In addition, we use the syntactic structure of the sentence to add new
  connections to the tree-structured decoder neural network (Seq2DRNN+SynC).
  We compare our NMT model with sequential and state of the art syntax-based
  NMT models and show that our model produces more fluent translations with
  better reordering.
  Since our model is capable of doing translation and constituency parsing at
  the same time we also compare our parsing accuracy against other neural
  parsing models.
\end{abstract}

\section{Introduction}
\label{sect:intro}
Neural machine translation (NMT) models were initially proposed as extensions
of sequential neural language models~\cite{sutskever2014sequence,Cho14,Bahdanau} or
convolutions over $n$-grams in the decoder~\cite{kalchbrenner2013recurrent}.
Early methods for discriminative training of machine translation models showed that the
loss functions for translation were not sensitive to the production of certain
important words such as verbs, without which the output sentence might be uninterpretable
by humans.
A good solution was to penalise such bad outputs using tree structures which
get very low scores if important words like verbs are
missing~\citep{chiang2005hierarchical,zollmann2006syntax,galley2006scalable}.
To this end, there has been a push to incorporate some syntax into NMT models:
\citet{Sennrich16} incorporate POS tags and dependency
information from the source side of a translation pair in NMT models.
\citet{Stahlberg16} use source language syntax to guide the
decoder of an NMT system to follow hierarchical structures (Hiero) rules
\citep{chiang2005hierarchical}.
\citet{Eriguchi16} and \citet{bastings2017graph} use tree-structured
encoders to exploit source language syntax.
\citet{aharoni2017towards} take the approach of serialising the parse trees
to use in a sequential decoder.
\citet{Eriguchi17} propose an NMT+RNNG model, which explores the possibilities
of using dependency syntax trees from the target language using StackLSTMs
\citep{Dyer15, Dyer16} to aid a sequential decoder.
These approaches showed promising improvements in translation quality
but all the models in previous work, even the model in \citet{Eriguchi17} which uses RNNG, are bottom-up
tree structured decoders.

In contrast, we use a top-down tree-structured model
called DRNN (Doubly-Recurrent Neural Networks) first proposed by \citet{Alvarez17}
to model structural syntactic information for NMT.
We call our novel NMT model Seq2DRNN, using DRNNs as a tree-structured decoder
combined with a sequential encoder and a novel syntax-aware attention
model.

All the previous work in syntax-aware NMT mentioned above has focused on dependency
parsing as the syntactic model. In contrast, we wish to pursue phrase structure
(aka constituency) based syntax-based NMT. We provide some analysis that shows
that constituency information can help recover information in NMT decoding.

We perform extensive experiments comparing our model
against other state-of-the-art sequence to sequence and syntax-aware NMT models
and show that our model can improve translation quality and reordering quality.
The model performs translation and constituency parsing simultaneously so we
also compare our parsing accuracy to other neural parsing models.


\section{Model Description}
\label{sect:model}
In this paper, source sentence will be written as
$\textbf{f}=x_1, x_2, ..., x_n$, target sentence as
$\textbf{e}=y_1, y_2, ..., y_m$ where $y_j$ is a word.
Additionally, $p_k$ represents a non-terminal symbol (constituent, phrase)
in the target sentence constituency tree.
$[\textbf{v}; \textbf{u}]$ stands for concatenation of vectors $\textbf{v}$ and
$\textbf{u}$. $\textbf{W}(x)$ is the word embedding of word $x$.

The design of our NMT system follows the encoder-decoder model (also known as a
sequence to sequence model) proposed by \citet{Cho14} and
\citet{sutskever2014sequence}.
Our system uses a standard bidirectional gated RNN (BiLSTM or bidirectional
Long Short-Term Memory) \citep{DBLP:journals/corr/HuangXY15} as the encoder
and our proposed tree-structured RNN as the decoder.

\subsection{Sequence to Sequence NMT (Seq2Seq)}
Neural machine translation models generally consist of an encoder, a decoder
and an attention model \citep{Luong15,Cho15}.
The encoder is used to produce hidden representations of the source sentence,
which is fed into the decoder along with the attention information to produce
the translation output sequence.

A common approach is to use bidirectional LSTMs as encoder, produce forward
hidden states $\overrightarrow{\textbf{h}}_i$ and backward hidden states
$\overleftarrow{\textbf{h}}_i$, and the final representation‚
$\textbf{h}^{\enc}_i$ is the concatenation of both:

\begin{equation}
\begin{aligned}
\overrightarrow{\textbf{h}}_i &=
    \overrightarrow{\textrm{RNN}}_{\enc}(
        \overrightarrow{\textbf{h}}_{i-1}, \textbf{W}_x(x_i)) \\[0em]
\overleftarrow{\textbf{h}}_i &=
    \overleftarrow{\textrm{RNN}}_{\enc}(
        \overleftarrow{\textbf{h}}_{i+1}, \textbf{W}_x(x_i)) \\[0em]
%
\textbf{h}^{\enc}_i &=
    [\overrightarrow{\textbf{h}}_i; \overleftarrow{\textbf{h}}_i]
\end{aligned}
\end{equation}

The decoder takes the output of the encoder and generates a sequence in target
language.
The attention mechanism provides additional context vectors $\textbf{c}_j$
which is a weighted average contribution of each $\textbf{h}_i$ source
side encoding.
\begin{equation}
\begin{aligned}
\textbf{h}^{\dec}_j &=
    \textrm{RNN}_{\dec}(
        \textbf{h}^{\dec}_{j-1}, [\textbf{o}_{j-1}; \textbf{c}_j]
    ) \\[0em]
\textbf{o}_j &=
    \softmax(\textbf{U}\textbf{h}^{\dec}_j + \textbf{b})
\end{aligned}
\end{equation}

Here, $\textbf{U}$ is the readout matrix and $\textbf{b}$ is the bias vector.
$\textbf{o}_j$ is the output word embedding.

\subsection{NMT with a Tree-Structured Decoder (Seq2DRNN)}
The output translation from a translation system should convey the same meaning
as the input.
This includes the correct word choices but also the right information
structure.
Sentence structure can be viewed as starting with an action or state (described
via verbs or other predicates) and the entities or propositions involved in
that activity or state (usually described via arguments to verbs).
Thus certain words in the output translation, like verbs, are crucial to the
understanding of the target language sentence but only provide marginal value
in $n$-gram matching evaluations like the BLEU score.
Tree representations, produced via dependency parsing and constituency parsing,
are useful because they are sensitive to this information structure.
Our tree-structured decoder uses a neural network to generate trees (described
in \S\ref{sssec:DRNN}), which is incorporated into an NMT model (our novel
encoder-decoder model is in \S\ref{sssec:Seq2DRNN}) which translates and
produces a parse tree.
Our new Syntactic Connection method (SynC) is described in \S\ref{sssec:SynC}
which is combined with the Seq2DRNN model (Seq2DRNN+SynC) and the attention
mechanism (\S\ref{sssec:Attention}).

\subsubsection{Doubly-Recurrent Neural Network}
\label{sssec:DRNN}
The Doubly-Recurrent Neural Network model \citep{Alvarez17} takes a vector
representation as input and generates a tree.
\citet{Alvarez17} show that the DRNN model can effectively reconstruct trees
but they do not use DRNNs within a full-scale NMT system. We also use DRNNs
for phrase-structure (aka constituency) tree structures rather than
dependency trees as in previous work.
DRNN decoding proceeds top-down; the generation of nodes at depth $d$ depend
solely on the state of nodes at depth $<d$.
Unlike previous work in tree-structured decoding for NMT by \citet{Dyer16}
and \citet{Eriguchi17}, the output sentence generation is not done in
sequence, where the target word $y_j$ is generated after all $y_{<j}$ are
generated.
DRNN first predicts the structure of the sentence and then expands each
component to predict words.
When generating $y_j$, information regarding the structure of words from $1$ to
$j-1$ and $j+1$ to $m$ can be used to aid prediction of $y_j$.

A DRNN consists of two recurrent neural network units, which separately process
ancestral and fraternal information about nodes in the tree.
Assuming a node is $v$, its immediate parent node is $P(v)$ and its closest
sibling on the left side (appears in the target language sequence just before
$v$) is $S(v)$.
The label of node $v$ is $\textbf{z}_v$.
Then the ancestral hidden representation $\textbf{h}^a$ and fraternal
representation $\textbf{h}^f$ of a node are calculated with
Equation~\ref{equ:DRNN_Node}.

\begin{equation}
\label{equ:DRNN_Node}
\begin{aligned}
\textbf{h}^a_v = \textrm{RNN}_{\dec}^a(\textbf{h}^a_{P(v)}, \textbf{z}_{P(v)})\\[0em]
\textbf{h}^f_v = \textrm{RNN}_{\dec}^f(\textbf{h}^f_{S(v)}, \textbf{z}_{S(v)})
\end{aligned}
\end{equation}

$\textbf{h}^a_v$ and $\textbf{h}^f_v$ are then combined to produce the hidden
state of node $v$ for prediction (predictive hidden state $\textbf{h}_v$,
Equation~\ref{equ:DRNN_PHS}),
which is used to predict the labels of node $v$.

\begin{equation}
\label{equ:DRNN_PHS}
\textbf{h}_v =
    \tanh(\textbf{U}^f\textbf{h}^f_v + \textbf{U}^a\textbf{h}^a_v)
\end{equation}

During the label prediction, DRNN first makes topological decisions: whether
(i) the current node is a leaf node (node with no children, $\alpha_v$);
then (ii) whether the current node has siblings on its right-hand side
($\gamma_v$). Both predictions are done using sigmoid activations:

\begin{equation}
\label{equ:DRNN_SigA}
\begin{aligned}
    o^a_v &= \sigma(\textbf{u}^a\textbf{h}_v) \\
    \alpha_v &= 1 \mbox{ if } o^a_v  \mbox{ is activated.}
\end{aligned}
\end{equation}

\begin{equation}
\label{equ:DRNN_SigF}
\begin{aligned}
    o^f_v &= \sigma(\textbf{u}^f\textbf{h}_v) \\
    \gamma_v &= 1 \mbox{ if } o^f_v  \mbox{ is activated.}
\end{aligned}
\end{equation}

Then, label representation $\textbf{o}_v$ is predicted using $\alpha_v$ and
$\gamma_v$, and the predictive hidden state $\textbf{h}_v$:
\begin{equation}
\label{equ:DRNN_Output}
\textbf{o}_v = \softmax(
    \textbf{U}_o\textbf{h}_v + \alpha_v\textbf{u}_a + \gamma_v\textbf{u}_f)
\end{equation}

At inference time each node at the same depth is expanded
independently, therefore the whole process can be parallelised.
This parallelism advantage is not observed in any of the sequential decoders
that generate output sequence strictly from left to right nor from right to
left (\S\ref{sect:rel_works} has more discussion).

\subsubsection{Parsing and Translating with DRNN}
\label{sssec:Seq2DRNN}
A DRNN is capable of producing a tree structure with labels given an input
vector representation.
If we train the DRNN to produce parse trees from the output of an encoder RNN,
this system will be able to translate and parse at the same time.
\citep{Alvarez17} in their paper provided a proof-of-concept NMT experiment
using dependency trees.
Instead of an single RNN unit to process fraternal information they had
multiple for modelling the fraternal information on syntactic tree structure
because dependency parse trees differentiate between left and right children.
The model itself also disregards the sequentiality of natural language, and
lack attention mechanisms to make it work in exchange for a strict top-bottom
decoding procedure.

\begin{figure}[t]
    \centering
    \includegraphics[scale=0.35]{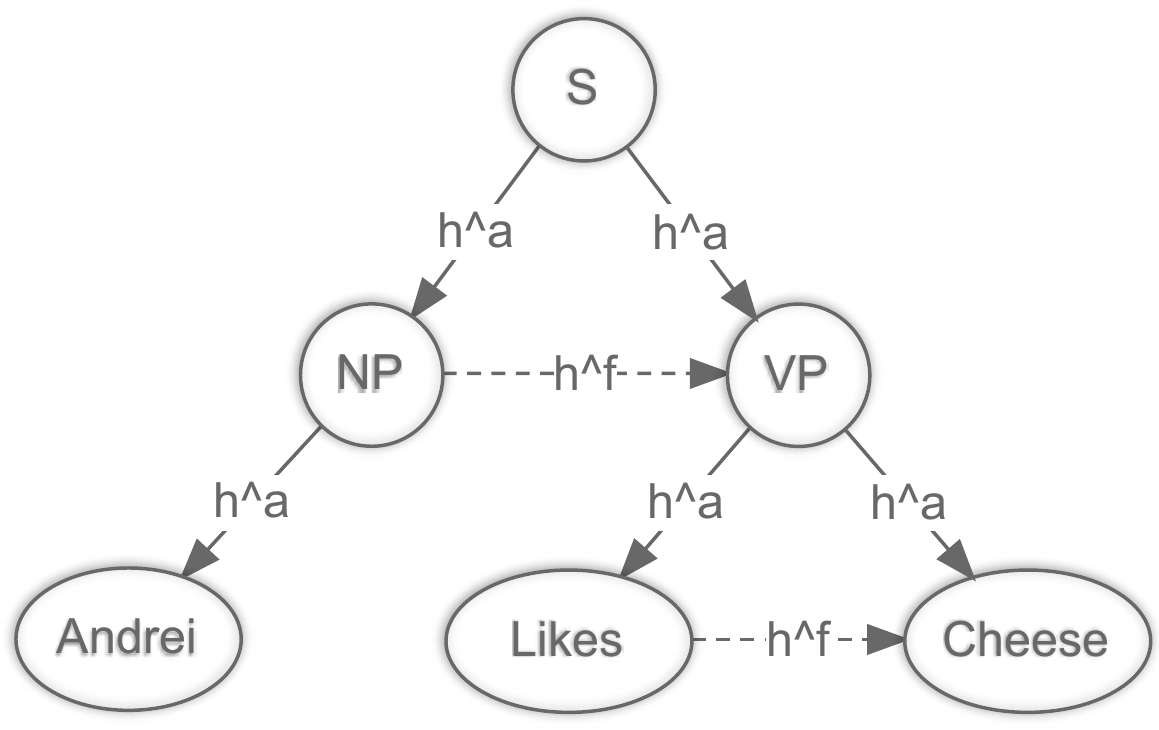}
    \caption{Seq2DRNN on Constituency Tree}\label{fig:DRNN}
\end{figure}

We use constituency parse trees to represent sentences in the target language
(Figure~\ref{fig:DRNN}) because constituency or phrase-structure trees
are more amenable to top-down derivation compared to dependency trees.
It is also easier to model for DRNN, and presumably more capable at handling
unknown words which is common in NMT systems with limited vocabulary size.

Each node on the tree represents either a terminal symbol (a word) or a
non-terminal symbol (a clause or phrase type).
The sub-tree dominated by a non-terminal node is the clause or phrase
identified with this non-terminal node label.

\begin{figure}[t]
    \centering
    \includegraphics[scale=0.4]{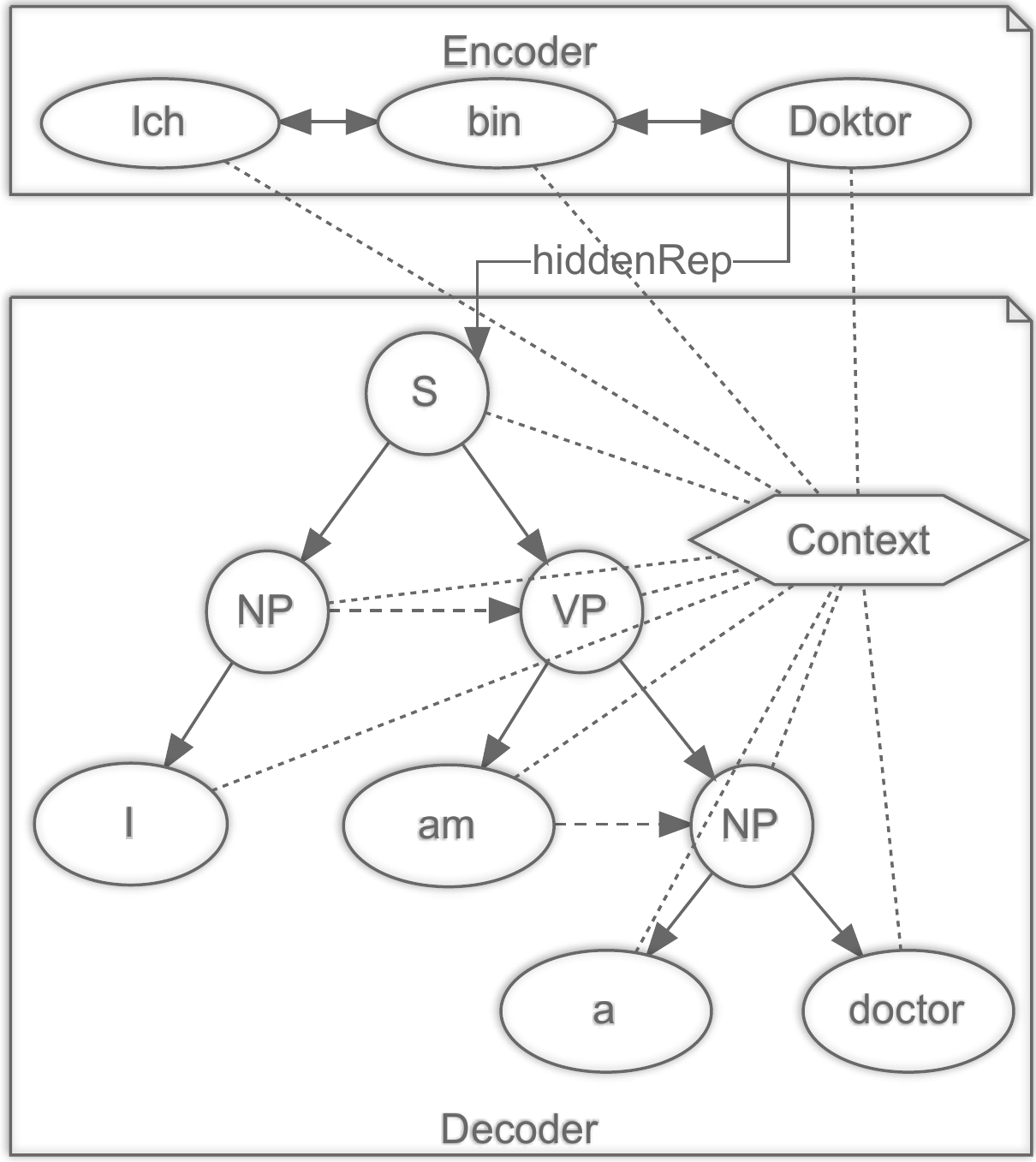}
    \caption{Seq2DRNN Encoder-Decoder}\label{fig:NMT-DRNN}
\end{figure}

A conventional bidirectional RNN (BiLSTM) encoder
\citep{Cho14,sutskever2014sequence} is used to produce hidden states for the
decoder (see Figure~\ref{fig:NMT-DRNN}).

We use breadth-first search to implement the Seq2DRNN decoder.
Two queues are used here: \textit{current} queue which is the queue containing
all of the nodes on the currently being processed depth, and \textit{next}
queue with nodes on the next depth (Algorithm~\ref{code:DRNN-Decode} has all
the details).

The decoding process starts from top to bottom, from root to its children, then
to its grandchildren, and so on until the leaf nodes which are the output
words.

In our implementation, sentence clauses (S nodes) are generated as the children
of the root node to generalise over sentence types and in case there are
multiple sentences in a single translation pair.

Initially, the \textit{current} queue will only have one entry: the root node, which
is initialised with the hidden representation of the source sentence.

Each node in the \textit{current} queue is expanded in the following manner: first
generate all of its siblings and add them to the \textit{current} queue, and if any
node happens to be non-terminal, generate its first child and add it to the
\textit{next} queue.
After the \textit{current} queue is empty, make \textit{next} the new \textit{current} queue and
start working on nodes at the next depth.

For training, we use back-propagation through trees using the approach in
\citet{Goller96}.
In the forward pass, the source sentence is encoded into a hidden
representation and fed into the decoder.
The decoder generates the tree, predicts the labels of every node from root to
leaves.
Then in the backward pass, gradients are calculated and used to update the
parameters.
The loss calculation includes losses in topological predictions: $o^a_v$ and
$o^f_v$ (Equations~\ref{equ:DRNN_SigA} and \ref{equ:DRNN_SigF}) and
label predictions: $\textbf{o}_v$ (Equation \ref{equ:DRNN_Output}).

\begin{equation}
\begin{aligned}
    \loss(\mbox{e}) =
    & \sum_{v}\loss^{\labels}(\textbf{o}_v, \hat{\textbf{o}_v}) + \\
    & \alpha \sum_{v}\loss^{\topo}(o^a_v, \hat{o^a_v}) + \\
    & \alpha \sum_{v}\loss^{\topo}(o^f_v, \hat{o^f_v})
\end{aligned}
\end{equation}

Here $\alpha$ is a hyper-parameter.
\begin{algorithm}[t]
\small
\caption{Seq2DRNN Decoder}\label{code:DRNN-Decode}
\begin{algorithmic}[1]
\Procedure{Decode(\textit{hiddenRep})}{}

\State $\textit{currentQueue} \gets \text{Node from }\textit{hiddenRep}$
\State $\textit{nextQueue} \gets \text{empty}$

\BState \emph{loop}:
\If {$\textit{currentQueue}\text{ is not empty}$}
    \State $\textit{node} \gets \textit{currentQueue.pop()}$
    \State $\textbf{Generate labels of }\textit{node}$
    \If {$\textit{node} \text{ has siblings}$}
        \State $\textit{currentQueue} \gets \text{sibling}(\textit{node})$
    \EndIf
    \If {$\textit{node} \text{ has children}$}
        \State $\textit{nextQueue} \gets \text{child}(\textit{node})$
    \EndIf
    \State \textbf{goto} \emph{loop}
\EndIf{}
\LeftComment{all nodes at current depth are generated}
\LeftComment{move on to the next depth}
\If {$\textit{nextQueue}\text{ is not empty}$}
    \State $\textit{currentQueue} \gets \textit{nextQueue}$
    \State $\textit{nextQueue} \gets \text{empty}$
    \State \textbf{goto} \emph{loop}.
\EndIf
\LeftComment{both queues should be empty now}
\EndProcedure
\end{algorithmic}
\end{algorithm}

\subsubsection{Attention Mechanism}
\label{sssec:Attention}
Attention mechanisms usually work by adding an additional context vector during
label prediction.
We use a variation of an existing attention mechanism proposed by
\citet{Luong15}.
In our attention model, we produce a context vector $\textbf{c}_v$ for every
node $v$ by looking at all hidden states produced by the encoder
$\textbf{h}^{\enc}_i$, then calculating the weights and adding up the weighted
hidden states.

\begin{equation}
\label{equ:attention_weight}
\mbox{weight}_{v,i} = \textbf{V}_a \tanh(
    \textbf{W}_a  \textbf{h}_v + \textbf{U}_a \textbf{h}^{\enc}_i)
    \in \mathbb{R}
\end{equation}
\begin{equation}
\textbf{c}_v = \sum^{n}_{i=1} \hat{\mbox{weight}_{v,i}} \textbf{h}^{\enc}_i
\end{equation}

After the calculation in Equation~\ref{equ:attention_weight}, the weights are
normalised with a softmax function before being used to calculate the context
vector.
The attention module allows the generation of labels to pay more attention to
specific token representations of words in the input sentence.

\subsubsection{SynC: Syntactic Connections for Language Generation
               (Seq2DRNN+SynC)}
\label{sssec:SynC}
A conventional Seq2Seq model uses an RNN language model
\citep{Cho14,sutskever2014sequence} conditioned on the input representation
produced by the encoder to generate the output one word at a time
(Equation~\ref{equ:LSTMLM}).
The prediction of a word $y_j$ is directly conditioned on previously
generated words where $\textbf{c}_j$ is the context vector.

\begin{equation}
\label{equ:LSTMLM}
    \pr(\textbf{e}) = \prod_j \pr(y_j| y_{<j}, \textbf{c}_j)
\end{equation}

The problem with this word-level language model is that it treats a sentence
as a plain sequence of symbols regardless of its syntactic construction.
Sentences may contain multiple subordinate clauses and their boundaries are
not well-modelled by sequential language models.

We propose a new method to connect the hidden units in the Seq2DRNN decoder
that pays attention to contextual tree relationships.
The prediction of the representation of a word or a constituent $z_j$ (if a
constituent then $p_j$, if a word then $y_j$) is defined as follows:

\begin{equation}
\label{equ:SynC}
\begin{aligned}
    \pr(z_j & \mid y_{<j}, \textbf{c}_j) = \\
        & \pr(z_j \mid y_{<j},
        y_k(\forall k, z_j \in p_k \lor \\
        & \precedes(y_k, z_j)) \mid \textbf{c}_j)
\end{aligned}
\end{equation}

The generation of the representation of a word/constituent $z_j$, which is part
of the clause that contain it ($z_j \in p_k$), with clauses before which
($\precedes(p_k, z_j)$), is conditioned on the following information:
i)  Word-level: previously generated words $y_{<j}$;
ii) Ancestral Clause: the clauses that contain the current word
        $p_k$, i.e.\ $(\forall k z_j \in p_k)$;
iii) Fraternal Clause: the clauses that precede the current clause
        $p_k$, i.e.\ $(\forall k \precedes(p_k, z_j))$.

In practice, the generation of a node looks at the following representations:

\begin{enumerate}
    \item Word-level: an RNN unit that produces the representation of previous
        words as a sequence $y_{<j}$;
    \item Ancestral: treating the ancestors of the current node as a sequence
        (from root to the immediate parent), the representation of that
        sequence: $p_k(\forall k, z_j \in p_k)$;
    \item Fraternal: treating the previous siblings of the current node as
        well as the previous siblings of its parent node and so on as a
        sequence, the representation of that sequence:
        $p_k(\forall k, \precedes(p_k, z_j))$.
\end{enumerate}

\begin{figure}
    \centering
    \includegraphics[scale=0.4]{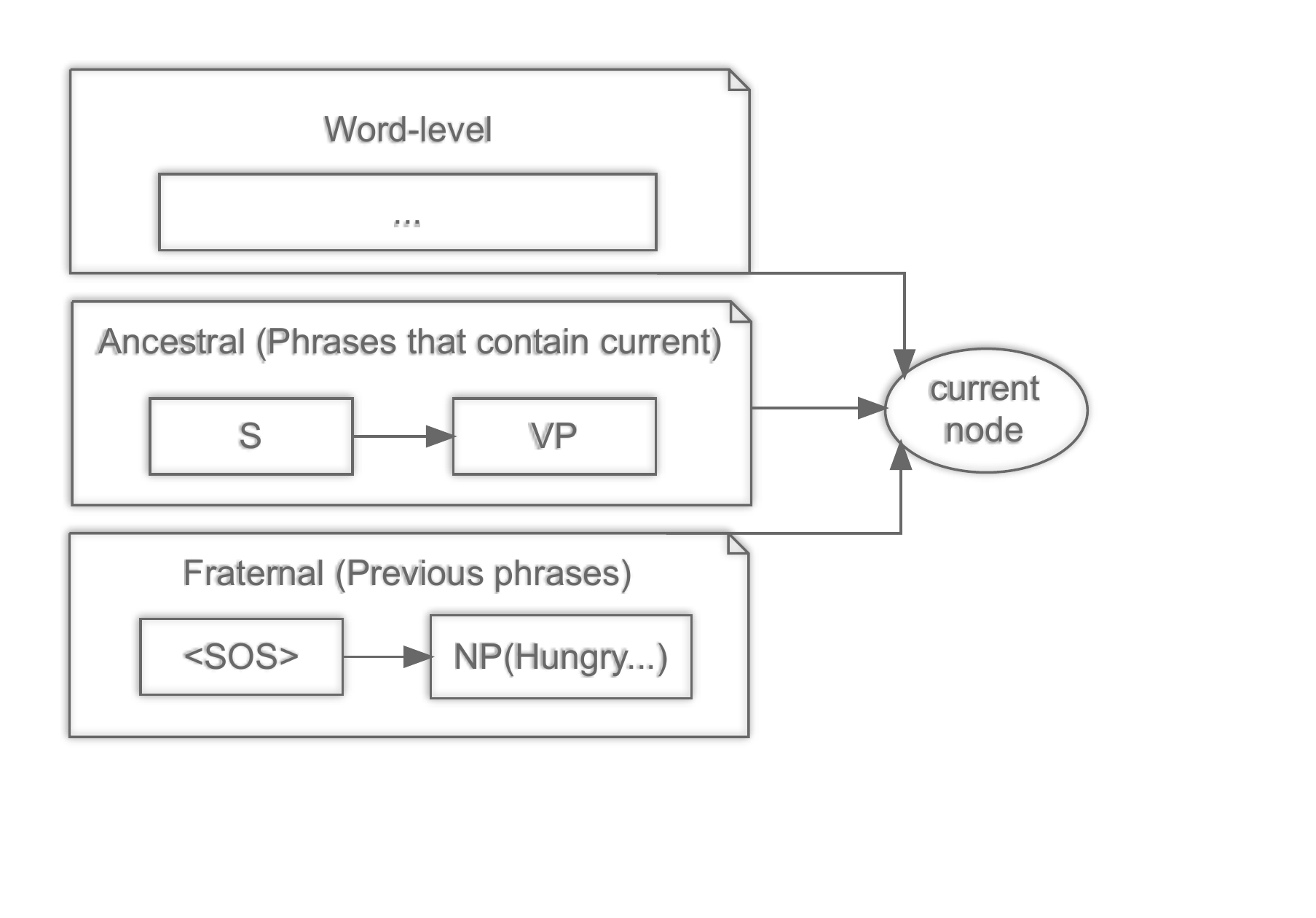}
    \caption{SynC example:
        the three types of information explicitly modelled when
        generating the current node/word \textit{likes} in
        \textit{Andrei when starving likes cheese}.
    }
    \label{fig:NMT-SynC}
\end{figure}

SynC creates connections in the tree-structured decoder that pays attention
to the structural context of generation of each terminal or non-terminal
symbol in the phrase structure tree.
For example in English, it is common for verb phrases to follow a noun phrase.
But that noun phrase could itself be a subordinate clause with its own
verb phrases.
In this case, our goal is to explicitly model the fact that the previous phrase
is a noun phrase instead of just the entire sequence of words.

\begin{figure}
    \centering
    \includegraphics[scale=0.4]{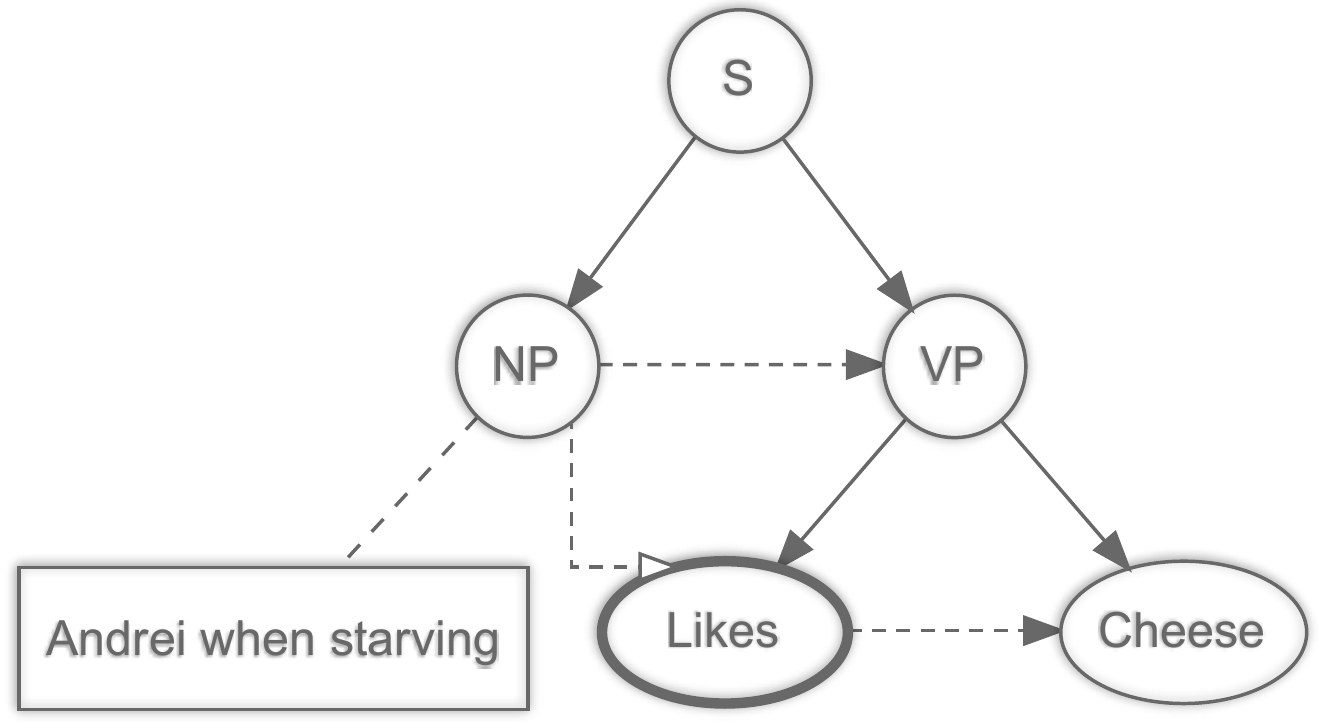}
    \caption{SynC in action;
        when generating the word \textit{likes} in
        \textit{Andrei when starving likes cheese},
        the prediction will be made knowing that the preceding clause is a
        noun phrase.}
    \label{fig:NMT-DRNN-SynC}
\end{figure}

SynC can be easily incorporated in the proposed Seq2DRNN model (Seq2DRNN+SynC).
In addition to the fraternal RNN unit that focuses on preceding sibling nodes,
and the ancestral DRNN unit that focus on parent nodes, a node would also look
at its parent's previous sibling state (the hidden vector representation of
preceding clauses from the very beginning of the sentence).
When a non-terminal symbol is expanded into a sub-tree, it's first child will
not have a previous sibling to provide fraternal information ($S(v) = \nan$,
as in Equ~\ref{equ:DRNN_Node}).
In this case, SynC establishes connection between its first child and its
parent's fraternal information provider for such fraternal RNN state
($S(v) = S(P(v))$).
\begin{equation}
\label{equ:DRNN_SynC}
\textbf{h}^f_v =
    \begin{cases}
        S(v) \neq \nan, & \textrm{RNN}_{\dec}^f(\textbf{h}^f_{S(v)}, \textbf{z}_{S(v)}) \\
        S(v) = \nan, & S(v) := S(P(v)),\\
                      &\textrm{RNN}_{\dec}^f(\textbf{h}^f_{S(v)}, \textbf{z}_{S(v)})
    \end{cases}
\end{equation}
An example is shown in Figure~\ref{fig:NMT-DRNN-SynC}.
In this case, a word-level language model will regard \textit{starving} as the
previous word, which is less helpful for the prediction of a verb phrase
\textit{likes cheese}.

\section{Experiments}
\label{sect:experiment}

\subsection{Model Training}
Experiments in this paper utilise constituency trees on the target side, these
trees are obtained by using the Stanford Lexical Parser \citep{Klein03} which we
chose for its speed and accuracy prior to training.

This procedure of pre-parsing data is not required at test time, our NMT system
would take a sentence as input and produces the translation in target language
along with its constituency tree as output.

We use the German-English dataset from IWSLT2017
\footnote{The International Workshop on Spoken Language Translation Evaluation
2017: \href{https://sites.google.com/site/iwsltevaluation2017/TED-tasks}{
https://sites.google.com/site/iwsltevaluation2017/TED-tasks}}
for our experiments, and tst2010-2015 as the test set
(Table~\ref{table:datainfo}).

\begin{table}[t!]
\small
    \begin{center}
        \begin{tabular}{|l|l|}
            \hline
            Train pairs & 226,572 \\ \hline
            Test pairs & 8,079 \\ \hline
            Unique source words & 128,857 \\ \hline
            Unique target words & 61,566 \\ \hline
            Average source sentence length & 21 \\ \hline
            Average target sentence length & 20 \\ \hline
        \end{tabular}
    \end{center}
    \caption{\label{table:datainfo} IWSLT2017 Dataset information}
\end{table}

To compare with other decoders that utilise target-side syntactic information,
we also evaluate on three more datasets from News Commentary v8 using
newstest2016 as testset (Table~\ref{table:NCDatainfo}).

\begin{table}[t!]
\small
    \begin{center}
        \begin{tabular}{|l|l|l|l|}
            \hline
            Language pair & DE-EN     & CS-EN     & RU-EN     \\ \hline
            Train pairs   & 166,313   & 134,453   & 131,492   \\ \hline
            Test  pairs   & 2,999     & 2,999     & 2,998     \\ \hline
            Uniq. src lex & 149,318   & 153,173   & 159,074   \\ \hline
            Uniq. tgt lex & 68,415    & 59,909    & 64,220    \\ \hline
            Avg. src len  & 25        & 22        & 25        \\ \hline
            Avg. tgt len  & 25        & 25        & 26        \\ \hline
        \end{tabular}
    \end{center}
    \caption{\label{table:NCDatainfo} News Commentary v8 Dataset information}
\end{table}

We replace all rarely occurring words with UNK (Unknown) tokens.
Only the top 50,000 most frequent words are kept.

\subsection{Modelling details}
The implementation of all models in this paper is done using DyNet
\citep{dynet} with Autobatching \citep{NeubigGD17}.
We use Long Short-Term Memory (LSTM) \citep{Hochreiter97} as RNN units.
Each LSTM unit has 2 layers, with input and hidden dimension of 256.
We use a minibatches of 64 samples.
\ifLossCurve
\else
We use early stopping mechanism for all experiments and Adam optimiser
\citep{kingma2014adam} as trainer.
\fi

Note that this configuration is significantly smaller in both dimension and batch
size than those presented in IWSLT2017 due to hardware limitations.
All experiments are carried out on a single GTX 1080 Ti GPU with 11GB of VRAM.

\subsection{Results}
\label{ssec:result}

\begin{table}[t!]
\small
    \begin{center}
        \begin{tabular}{|l|l|l|l|l|}
            \hline
                        & BLEU   & RIBES    & Perplx. \\ \hline\hline
            Seq2Seq     & 22.83  & 81.5     & 1.828      \\ \hline
            Seq2DRNN    & 23.53  & 80.4     & \textbf{1.644}      \\ \hline
            Seq2DRNN+SynC & \textbf{25.36}
                        & \textbf{82.6}
                        & 1.750  \\ \hline
        \end{tabular}
    \end{center}
    \caption{\label{table:result} IWSLT17 Experiment results}
\end{table}

\begin{table*}[t]
\small
\centering
    \begin{tabular}{|l|l|l|l|l|l|l|}
        \hline
        Dataset     & \multicolumn{2}{c|}{DE-EN}
                    & \multicolumn{2}{c|}{CS-EN}
                    & \multicolumn{2}{c|}{RU-EN} \\\hline
                    & BLEU  & RIBES & BLEU  & RIBES & BLEU  & RIBES \\\hline
        \hline
        Seq2Seq     & 16.61 & 73.8  & 11.22 & 69.6  & 12.03 & 69.6  \\\hline
        Str2Tree    & 16.13 & ---   & 11.65 & ---   & 11.94 & ---   \\\hline
        NMT+RNNG    & 16.41 & 75.0  & 12.06 & \textbf{70.4}
                                                    & 12.46 & 71.0  \\\hline
        \hline
        Seq2DRNN    & 16.90	& 75.1  & 11.84 & 67.3  & 12.04 & 69.7  \\\hline
        Seq2DRNN+SynC & \textbf{17.21} & \textbf{75.8}
                    & \textbf{12.11} & 70.3
                    & \textbf{12.96} & \textbf{71.1}  \\\hline
    \end{tabular}
    \caption{
        \label{table:NCresult} News Commentary v8 Experiment results.
        Seq2Seq and NMT+RNNG results are taken from \citet{Eriguchi17},
        Str2Tree (string-to-linearised-tree) results (no RIBES scores) come
        from \citet{aharoni2017towards}
        All numbers reported here are of non-ensemble models.
    }
\end{table*}

Table~\ref{table:result} and Table~\ref{table:NCresult} has the BLEU
\citep{Papineni02} and RIBES \citep{Isozaki10} scores.
In our IWSLT2017 tests, both Seq2DRNN and Seq2DRNN+SynC produce better results
than the Seq2Seq baseline model in terms of BLEU scores, while Seq2DRNN+SynC also
produces better RIBES scores indicating better reordering of phrases in the
output.
The Seq2DRNN+SynC model performs better than the Seq2DRNN model.
Both Seq2Seq and Seq2DRNN+SynC are able to produce results
with lower perplexities than the baseline Seq2Seq model on the test data.

In our News Commentary v8 tests, the same relative performance from Seq2DRNN(SynC)
can be observed.
The Seq2DRNN+SynC model is also able to out-perform the Str2Tree
model proposed by \citet{aharoni2017towards} and NMT+RNNG by \citet{Eriguchi17}
in most cases.
Note that \citet{Eriguchi17} used dependency information instead of constituency
information as presented in our work and \citet{aharoni2017towards}'s work.

\begin{table*}[t]
\small
\centering
    \begin{tabular}{|l|p{12cm}|}
    \hline
    Source  & wir wiederholten diese \"ubung mit denselben studenten. was
            glauben sie passiert nun? nun verstanden sie den vorteil des
            prototyping. so wurde aus demselben, schlechten team eines unter
            den besten. sie produzierten die h\"ochste konstruktion in der
            geringsten zeit.
            \\\hline
    Literal & we repeated this exercise with the same students. now what do you
            believe happened? now they understand the value of prototyping. so
            the same terrible team became one of the very best. they produced
            the tallest construction in the shortest time.\\ \hline
    Gold    & we did the exercise again with the same students. what do you
            think happened then? so now they understand the value of
            prototyping. so the same team went from being the very worst to
            being among the very best. they produced the tallest structures in
            the least amount of time. \\ \hline
    Seq2Seq & we repeated this with the same students. what happened you think
            differently? now, you know, the advantage of the design of the
            cycle. so, the same one of the team of the team among the best. it
            produced songs in the slightest building. \\ \hline
    Seq2DRNN & we’ll repeat these queries with the same students. what do you
            think of this? now it understood the advantage of the interests.
            that's been made of the same thing of one thing. they produced the
            highest construction of the best time at the best time. \\ \hline
    Seq2DRNN+SynC & we repeated this practice with the same students. what do you
            think happened? now, they understood the value of prototyping. it
            was being made of the same thing of one of the best ones. they
            produced the highest construction in the best time. \\ \hline
    \end{tabular}

\caption{\label{table:translation_sample}
    Translation Sample.
    \textit{Gold} is the reference, and \textit{Literal} is produced by a
    bilingual German-English speaker.
}
\end{table*}

Table~\ref{table:translation_sample} shows an example translation from all of
the models we use in our experiments.
Seq2Seq is able to translate with the correct vocabulary, but the sentences are
often syntactically awkward.
As the sentence length increases the syntactic fluency of Seq2Seq gets worse.
Seq2DRNN is able to produce more syntactically fluent sentences since each
lowest sub-clause contains typically $\leq 5$ words.
Seq2DRNN+SynC produces the best results in this example: produces more
syntactically fluent sentences, chooses the right words in the right place more
frequently.


We also took several examples from our IWSLT17 experiment and blank out certain
nouns by replacing them with unknown tokens
(Table~\ref{table:noun_unknown_sample}).
Note that in our training set, most sentences do not have unknown tokens, and
those that do only have at most 1.
Our assumptions of the observed patterns in this case are: i) the proposed
models are more capable at handling unknown tokens; ii) while Seq2DRNN is more
capable at retaining the structure of the sentence, it cannot rely on a wider
context to predict certain common phrases with noises in the source sentence;
iii) the proposed Seq2DRNN+SynC model is more capable at handling unknown words
both in the sense of being better at retaining sentence structure and handling
noisy input.

\begin{table*}[t]
\small
\centering
    \begin{tabular}{|l|l|p{6.5cm}||p{6.5cm}|}
    \hline
    0 &
    src     & es war \textbf{zeit} zum \textbf{abendessen} und wir hielten
            ausschau nach einem \textbf{restaurant}.
            & die gute \textbf{nachricht} ist , dass die \textbf{person} , die
            das gesagt hat \textbf{ann coulter} war . \\
    & S2S   & and it was \textbf{time} for \textbf{dinner} , and we were
            looking for a \textbf{restaurant}.
            & the good \textbf{news} is that the \textbf{person} who said that
            was \textbf{ann coulter} . \\
    & S2D   & it was the \textbf{time} for \textbf{dinner} , and we were
            looking for a \textbf{restaurant}.
            & the good \textbf{news} is that the \textbf{person} who said that
            was \textbf{ann coulter} . \\
    & S2D+L & it was \textbf{time} for \textbf{dinner} , and we were
            looking for a \textbf{restaurant}.
            & the good \textbf{news} is that the \textbf{person} who said
            that was \textbf{ann coulter} . \\ \hline

    1 &
    src     & es war \textbf{zeit} zum \textbf{abendessen} und
            wir hielten ausschau nach einem \textbf{UNK}.
            & die gute \textbf{UNK} ist , dass die \textbf{person} , die
            das gesagt hat \textbf{ann coulter} war . \\
    & S2S   & and it was \textbf{time} for \textbf{dinner} , and we \underline{
            thought we} \underline{were looking for a window}.
            & the good \textbf{news} is that the \textbf{person} who said that
            \underline{the teacher} was \textbf{ann coulter} . \\
    & S2D   & it was the \textbf{time} for \textbf{dinner} ,
            and we were looking for a \textbf{UNK} \underline{pilot}.
            & the good \underline{motivator} is that the \textbf{person} who
            said that was \textbf{ann coulter} . \\
    & S2D+L & it was \textbf{time} for \textbf{dinner} , and we were
            looking for a \textbf{UNK}.
            & the good \textbf{news} is that the \textbf{person} who said
            that was \textbf{ann coulter} . \\ \hline

    2 &
    src     & es war \textbf{zeit} zum \textbf{UNK} und
            wir hielten ausschau nach einem \textbf{UNK}.
            & die gute \textbf{UNK} ist , dass die \textbf{UNK} , die
            das gesagt hat \textbf{ann coulter} war . \\
    & S2S   & and it was \textbf{time} for \underline{the time time} , and we
            \underline{thought} \underline{we looked at a window search}.
            & the good \textbf{news} is that the \textbf{UNK} that \underline{
            the UNK , which}
            was \textbf{ann coulter} . \\
    & S2D   & it was the \underline{second} \textbf{time} \underline{   },
            and we \underline{would look} for a \textbf{UNK}
            \underline{pilot}.
            & the good \underline{motivator} is that the \textbf{UNK group} who
            said that was \textbf{ann coulter}. \\
    & S2D+L & it was a \textbf{time} for \textbf{UNK} , and we were
            looking for a \textbf{UNK}.
            & the good \textbf{news} is that the \textbf{people} who said
            that was \textbf{ann coulter} . \\ \hline

    3 &
    src     & es war \textbf{UNK} zum \textbf{UNK} und
            wir hielten ausschau nach einem \textbf{UNK}.
            & die gute \textbf{UNK} ist , dass die \textbf{UNK} , die
            das gesagt hat \textbf{UNK} war . \\
    & S2S   & \underline{UNK} was a \underline{time time and UNK} ,
            \underline{looking for a} \underline{UNK look for a look at a war}.
            & \underline{UNK is not a cop 's needs to be ozzie good to good}
            \underline{good , which is that the UNK UNK that said that it}
            \underline{ was said} . \\
    & S2D   & it was \textbf{time} \underline{of time and guerrilla} .
            & the good \underline{motivator} is that the \textbf{UNK}
            \underline{planner} that said this was a \textbf{UNK}
            \underline{video}. \\
    & S2D+L & it was \textbf{time} for \underline{the tone} and
            \underline{night to watch the} \underline{connection} .
            & the good \textbf{news} is that the \textbf{people} that said
            that was \underline{mandated} . \\ \hline
    \end{tabular}

\caption{\label{table:noun_unknown_sample}
    Unknown noun experiment samples.
    Substituted and correct nouns are marked in \textbf{Bold}, while incorrect
    elements are marked in \underline{underline}.
    Examples shown are: no UNK; 1 UNK; 2 UNKs; 3 UNKs.
    When there are no unknown tokens, all three compared models are able to produce
    reasonably good if not identical translations.
    When there is only one UNK token, Seq2DRNN often does not use
    the context to predict an appropriate word or phrase. In contrast, both
    the Seq2Seq and Seq2DRNN+SynC  were able to correctly predict that
    \textit{die gute UNK ist} could be translated to \textit{the good news is}.
    When there are 2 UNK tokens in the source sentence, Seq2Seq produces more incorrect
    predictions, Seq2DRNN makes some mistakes, while Seq2DRNN+SynC is
    able to get the most parts correct.
    Finally, when we replace 3 nouns, all models fail to some degree
    while Seq2Seq's output is the worst.
}
\end{table*}

\subsection{Attention Module}
\begin{figure}[t]
    \centering
    \includegraphics[width=0.3\textwidth]{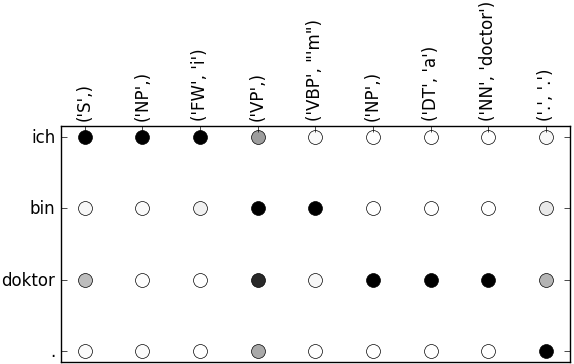}
    \caption{Attention Module in Seq2DRNN+SynC.
        ``ich" means ``I", ``bin" is ``am", ``doktor"'s literal translation is
        ``doctor".
        Darker colour means higher weight (relevance score) as calculated in
        Equation~\ref{equ:attention_weight}.
        The values of each column sum up to 1.
        The attention weights in this example perfectly align with the appropriate
        clauses.
        Additional example is provided in the appendix.
    }
    \label{fig:attention}
\end{figure}

We visualise the attention weights of our Seq2DRNN+SynC model.
Attention~\S\ref{sssec:Attention} computes a context vector for each node in
the tree (a weighted sum of the source side vector representations).
For the translation pair in Figure~\ref{fig:NMT-DRNN}, we show the attention
weight of each pair of word and node (Equation~\ref{equ:attention_weight}) in
Figure~\ref{fig:attention}.

The addition of syntax nodes in the output enables the attention model to be
used more effectively and is also valuable for visual inspection of syntactic
nodes in the output mapping to the input.

\subsection{Parsing Quality}
To evaluate the parsing quality, we follow the approach by \citet{Vinyals15}
and train a DRNN(SynC) model to produce English to English(Tree) translation.
We use the same data and experiment settings that \citet{Vinyals15} used:
the Wall Street Journal Penn Treebank English corpus with golden constituency
structure, 256 for input/hidden dimension and 3 layers of RNN.
We evaluate on section 23 of the aforementioned WSJ data using EVALB\footnote{
\href{https://nlp.cs.nyu.edu/evalb/}{https://nlp.cs.nyu.edu/evalb/}}.
The results are presented in Table~\ref{table:parser}.

\begin{table}[t!]
    \begin{center}
        \begin{tabular}{|l|l|}
            \hline
            Model                      & F-measure \\ \hline\hline
            Baseline \citep{Vinyals15} & $<$ 70    \\ \hline
            LSTM+AD \citep{Vinyals15}  & 88.3      \\ \hline\hline
            \citet{Petrov10}           & 91.8      \\ \hline
            \citet{Dyer16}             & \textbf{92.4}\\ \hline\hline
            Seq2DRNN                   & 89.4      \\ \hline
            Seq2DRNN+SynC                & 89.9      \\ \hline
        \end{tabular}
    \end{center}
    \caption{
        Parser scores.
        Numbers from \citep{Vinyals15} are of non-ensemble models.
    }
    \label{table:parser}
\end{table}

Although falling short behind more specifically designed models like the RNNG
by \citet{Dyer16}, our model is able to produce better results than the LSTM+AD
model proposed by \citet{Vinyals15}, which is more comparable to ours since
they are also using an NMT model to do constituency parsing.
Since our work is more focusing on the translation aspect, optimising and
designing a dedicated parser is slightly off-topic here.
Nevertheless, it is worth noting that in 50.89\% of the cases, Seq2DRNN+SynC
was able to produce output that perfectly matches the reference.
The same number for sentences with less than 40 words is 52.16\%, while the
F-measure increases to 90.5.
This shows Seq2DRNN(SynC) when doing parsing can produce outputs of similar
quality when handling longer sentences.

\ifParseIWSLT
We also do evaluation on our translation results from the IWSLT dataset.
Since translation results do not come with reference parse trees, we parse the
output of our decoder using the same parser we used in our other experiments:
the Stanford Parser.
Constituency parsing evaluation is done using Precision/Recall/F1-scores
on the output constituent spans (unlabelled) and spans and labels (labelled).
The results are presented in Table~\ref{table:unlabelled} and
Table~\ref{table:labelled}.
The parser we use gets F1 score of 87.04 on Penn Treebank English
constituency parsing \citep{Klein2003}.

\begin{table}[t!]
    \begin{center}
        \begin{tabular}{|l|l|l|l|l|}
            \hline
                    & \multicolumn{3}{c|}{Unlabelled} \\ \hline
                    & Prec.  & Rec.   & F1    \\ \hline\hline
            Seq2DRNN    & 96.87  & 96.93  & 96.90 \\ \hline
            Seq2DRNN+SynC & 96.43  & 95.89  & 96.16 \\ \hline
        \end{tabular}
    \end{center}
    \caption{IWSLT Translation result constituency unlabelled scores.
        Reference parse trees obtained using Stanford Parser.
    }
    \label{table:unlabelled}
\end{table}

\begin{table}[t!]
    \begin{center}
        \begin{tabular}{|l|l|l|l|l|}
            \hline
                    & \multicolumn{3}{c|}{Labelled} \\ \hline
                    & Prec.  & Rec.   & F1    \\ \hline\hline
            Seq2DRNN    & 91.63  & 91.69  & 91.66 \\ \hline
            Seq2DRNN+SynC & 90.73  & 90.22  & 90.48 \\ \hline
        \end{tabular}
    \end{center}
    \caption{IWSLT Translation result constituency labelled scores.
        Reference parse trees obtained using Stanford Parser.
    }
    \label{table:labelled}
\end{table}


The presence of SynC in the decoder influences parse tree construction:
the Seq2DRNN+SynC F1 score is comparable but lower than Seq2DRNN.
\fi

\section{Related Work}
\label{sect:rel_works}
Recent research shows that modelling syntax is useful for various neural
NLP tasks.
\citet{Dyer15,Dyer16,Vinyals15,luong2016multi} have works on
language modelling and parsing, \citet{tai2015improved} on semantic
analysis, and \citet{Zhang15} on sentence completion, etc.

\citet{Eriguchi17} showed that NMT model can benefit from neural syntactical
parsing models.
\citet{charniak2016parsing} showed that a neural parsing problem
shares similarity to neural language modelling problem, which forms a building
block of an NMT system.
We can then make the assumption that structural syntactic information utilised
in neural parsing models should be able to aid NMT, which is shown to be true
here.

\citet{Zhang15} proposed TreeLSTM which is another structured neural decoder.
TreeLSTM is not only structurally more complicated but also uses external
classifiers.
\citet{dong2016language} also proposed a sequence-to-tree (Seq2Tree) model
for question answering.
Both of these models are not designed for NMT and lack a language model.
While operate from top-to-bottom like Seq2DRNN(+SynC), TreeLSTM and Seq2Tree
produce components that lack sequential continuity which we have shown to be
non-negligible for language generation.

\citet{aharoni2017towards}, \citet{Wu17}, and \citet{Eriguchi17} experimented
with NMT models that utilise target side structural syntax.
\citet{aharoni2017towards} treated constituency trees as sequential strings
(linearised-tree) and trained a Seq2Seq model to produce such sequences.
\citet{Wu17} proposed SD-NMT, which models dependency syntax trees by adding a
shift-reduce neural parser to a standard RNN decoder.
\citet{Eriguchi17} in addition to \citet{Wu17}'s work, proposed NMT+RNNG which
uses a modified RNNG generator \citep{Dyer16} to process dependency instead of
constituency information as originally proposed by \citet{Dyer16}, making it
consequently a StackLSTM sequential decoder with additional RNN units so it
is still a bottom-up tree-structured decoder rather than a top-down decoder like ours.
Nevertheless, all of these research showed that target side syntax could
improve NMT systems. We believe these models could also be augmented with
SynC connections (with NMT+RNNG one has to instead use constituency information).

\section{Conclusions}
\label{sect:conc}

We propose an NMT model that utilises target side constituency syntax with
a strictly top-down tree-structured decoder using Doubly-Recurrent Neural
Networks (DRNN) incorporated into an encoder-decoder NMT model.
We propose a new way of modelling language generation by establishing
additional clause-based syntactic connections called SynC.
Our experiments show that our proposed models can outperform a strong
sequence to sequence NMT baseline and several rival models and do parsing
competitively.

In the future we hope to incorporate source side syntax into the model.
We plan to explore the applications of SynC in NMT with more structured
attention mechanisms, and potentially a hybrid phrase-based NMT systems with
SynC, in which the model can benefit from SynC to be more extensible when
handling larger lexicons.

\bibliography{acl2018}
\bibliographystyle{acl_natbib_nourl}

\clearpage
\appendix
\onecolumn

\section{Additional Translation Samples (IWSLT2017 German-English)}
The samples here are from our IWSLT2017 German-English testset.
We compared the performances of all our proposed models as well as the baseline
Seq2Seq model in Table~\ref{table:translation_sample_extra}.

\begin{figure*}[htb!]
    \centering
    \includegraphics[width=0.8\textwidth]{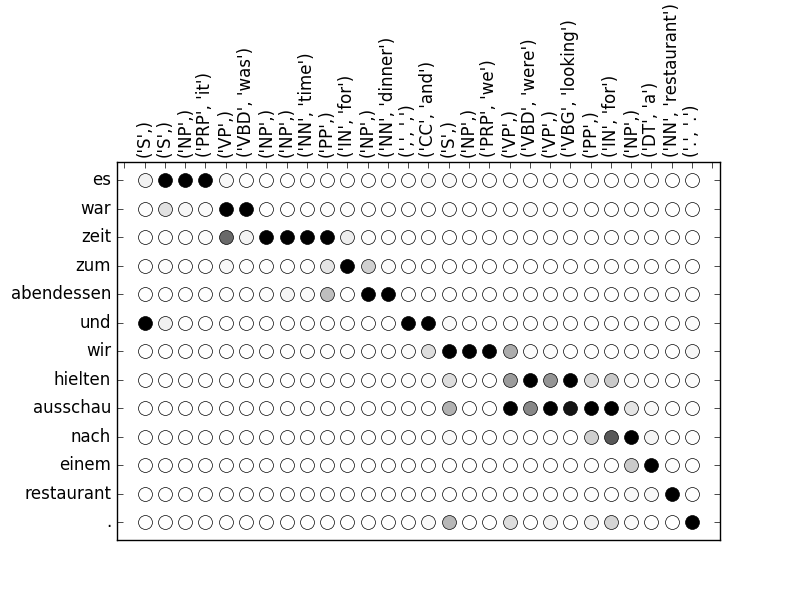}
    \caption{
        Attention visualisation (Seq2DRNN+SynC).
        Darker colour means higher attention weight as defined in
        Equ~\ref{equ:attention_weight}.
        The sentence is randomly selected from our IWSLT experiment.
    }
    \label{fig:attention_extra}
\end{figure*}

\begin{figure*}
    \centering
    \includegraphics[width=\textwidth]{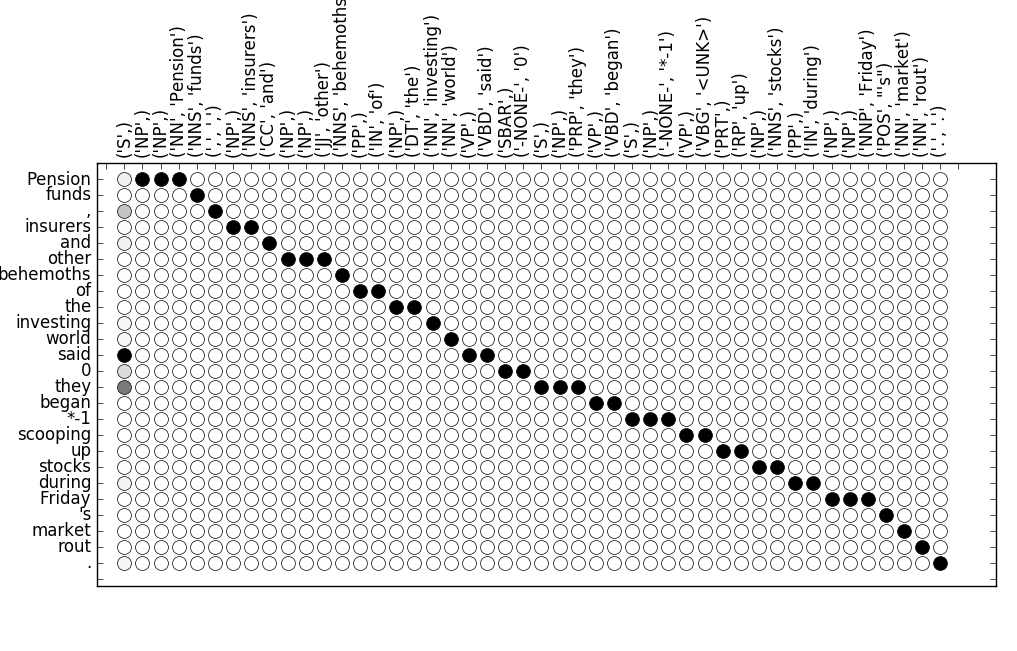}
    \caption{
        Parser attention visualisation (Seq2DRNN+SynC).
        Darker colour means higher attention weight as defined in
        Equ~\ref{equ:attention_weight}.
        The sentence is randomly selected from our PennTreebank experiment.
    }
    \label{fig:attention_parser}
\end{figure*}

We provide an additional example of our attention module visualisation in
Figure~\ref{fig:attention_extra} and for parser in
Figure~\ref{fig:attention_parser}.

\begin{table*}[htb!]
\centering
    \begin{tabular}{|l|p{12cm}|}
    \hline
    Source  & 1 m \underline{ist} das h\"ochste, was
            \underline{ich gesehen habe}.
            \\
    Literal & 1 m \underline{is} the highest that \underline{i've seen}.
            \\
    Reference
            & thirty-nine inches \underline{is} the tallest structure
            \underline{i've seen}.
            \\
    Seq2Seq & and the highest thing \textbf{\underline{is}
            \underline{i've seen}}.
            \\
    Seq2DRNN & one \underline{is} the highest thing \underline{i've seen}
            at that.
            \\
    Seq2DRNN+SynC
            & feet \underline{is} the highest thing \underline{i've seen}.
            \\ \hline
    Source  & ich wei{\ss} nicht . sie \underline{wollten in die zeit zurück} ,
            bevor \underline{es} autos \underline{gab}
            oder twitter oder amerika sucht den superstar.
            \\
    Literal & i don’t know. they \underline{want to go back in time}, before
            \underline{there were} automobiles or twitter or america looking
            for superstar.
            \\
    Reference
            & i don’t know. \underline{they want to go back} before
            \underline{there were} automobiles or twitter or american idol.
            \\
    Seq2Seq & i don't know. they \textbf{\underline{were in the days}}, when
            \textbf{\underline{they were}} cars before
            \textbf{cars or the earnings, or america, and the country}.
            \\
    Seq2DRNN & i don't want to know before time, they
            \underline{wanted to go back} before
            \textbf{the cars before \underline{they were} cars or americans}.
            \\
    Seq2DRNN+SynC
            & i don't know. they \underline{wanted to go back in time},
            \textbf{they wanted to go back into the before,}
            before \underline{there had} cars or twitter \textbf{visitors}.
            \\ \hline
    \end{tabular}

\caption{\label{table:translation_sample_extra}
    Translation Samples.
    \textit{Gold} is the reference, and \textit{Literal} is produced by a
    bilingual German-English speaker.
    The reason we include the literal translation is that sometimes the
    reference translation from the corpus can have additional components or be
    non-literal translations.
}
\end{table*}

\end{document}